# Generative Data Refinement: Just Ask for Better Data

**Minqi Jiang[2], João G. M. Araújo[1], Will Ellsworth[2], Sian Gooding[1] and Edward Grefenstette[1]**
[1]Google DeepMind, [2]Work done while at Google DeepMind

**For a fixed parameter size, the capabilities of large models are primarily determined by the quality and quantity of its training data. Consequently, training datasets now grow faster than the rate at which new data is indexed on the web, leading to projected data exhaustion over the next decade. Much more data exists as user-generated content that is not publicly indexed, but incorporating such data comes with considerable risks, such as leaking private information and other undesirable content. We introduce a framework, *Generative Data Refinement* (GDR), for using pretrained generative models to transform a dataset with undesirable content into a refined dataset that is more suitable for training. Our experiments show that GDR can outperform industry-grade solutions for dataset anonymization, as well as enable direct detoxification of highly unsafe datasets. Moreover, we show that by generating synthetic data that is conditioned on each example in the real dataset, GDR's refined outputs naturally match the diversity of web scale datasets, and thereby avoid the often challenging task of generating diverse synthetic data via model prompting. The simplicity and effectiveness of GDR make it a powerful tool for scaling up the total stock of training data for frontier models.**

## 1. Introduction

The predictable scaling of model performance as a function of both parameter size and amount of training data is one of the most consequential findings in large-scale generative modeling. Such scaling laws (Hoffmann et al., 2022; Kaplan et al., 2020) suggest that when increasing the FLOPs budget for training a transformer-based large-language model (LLMs), both model parameters and training tokens must be scaled proportionately to remain compute-optimal in achieving the best test loss. These findings have sparked a rapid scaling up of model parameter counts and training dataset sizes. As continued scaling of model sizes is often impractical for many use cases and organizations, there has been further intensified focus along data scaling. Consequently, training datasets are now estimated to be expanding faster than the rate at which new data is indexed on the web, leading to projected data exhaustion over the next decade (Villalobos et al., 2022).

This analysis, however, is based on the size of publicly indexed datasets. Much more data is created on a continual basis in the form of content that is not publicly indexed on the web (GSMA, 2022; Radicati Group., 2020). This content includes user-generated data and many other forms of proprietary information. Training on this category of data presents several crucial risks, notably the potential of models memorizing private information, toxic content, and copyrighted material. Perhaps, with these risks in mind, many recent data scaling efforts have focused on devising protocols for producing *synthetic data*—useful data outputs sampled directly from a pretrained model or a model finetuned on an exemplar dataset. Often samples can be further filtered against a proxy reward model that captures the target criteria. Such purely synthetic approaches carry their own additional costs and risks: Fine-tuning the model requires additional compute and serving overhead (Rafailov et al., 2024). Moreover, the process can overfit to the reward model (Gao et al., 2023) as well as collapse to a small subset of possible samples satisfying the target criteria (Kirk et al., 2023). Importantly, in many domains, synthetic samples will often appear markedly distinct from the natural data they seek to emulate.

We introduce a distinct problem framing for the task of synthetic data generation, called *Generative Data Refinement* (GDR). In GDR, we apply a pretrained generative model to modify a

*Corresponding author(s): etg@google.com*

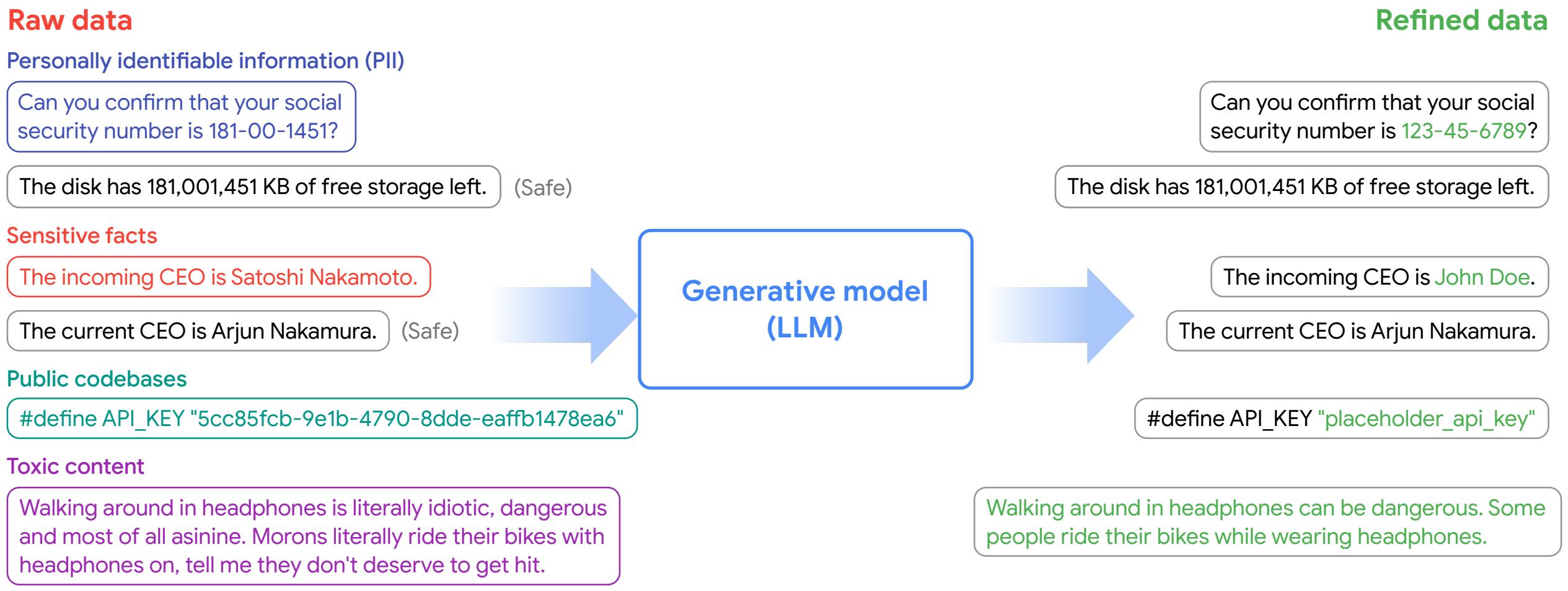

Figure 1 | Overview of Generative Data Refinement (GDR). Raw data, potentially containing undesirable content according to one or multiple criteria (e.g. personally identifiable information, toxic content) alongside information which should be retained, is passed to a pretrained generative model, e.g. a prompted LLM trained on large-scale web data, which uses its rich world knowledge to refine each sample to be free of the undesirable content while retaining any otherwise appropriate contents. Thus, GDR produces a refined dataset suitable for training.

collection of real data samples to be free of any undesirable content, such as potentially private facts, while preserving any otherwise useful information. We will refer to synthetic data that is anchored to real data as *grounded* synthetic data. By relying on the content of real data samples, GDR methods naturally produce more realistic outputs that also capture the diversity inherent in large-scale datasets, thereby side-stepping core issues with other synthetic data approaches. Moreover, GDR can be combined with additional methods for domain-specific adaptation of the model component, such as task-specific fine-tuning, many-shot prompting (Agarwal et al., 2024a; Anil et al., 2024), and search with a reward model.

In this work, we investigate the application of GDR to text and code data, with a focus on the common data sanitation tasks of anonymization and detoxification. Our experiments show that GDR, with either a zero-shot prompt or in combination with other methods like few-shot prompting and fine-tuning, can significantly improve over industry-standard methods for data anonymization in several data domains, including large-scale synthetic data domains and real-world codebases. Moreover, we find GDR is effective in cleaning highly toxic conversation logs, producing refined datasets with significantly-reduced degrees of toxicity. For both forms of data sanitization, we show the refined outputs of GDR can be used to train models that acquire useful information in the original dataset, without leaking private or toxic content in their outputs.

## 2. Related works

### 2.1. Synthetic data generation

Many recent data scaling efforts focus on devising protocols for generating high-quality samples from pretrained models, as aligned with a human preference model or some other reward model. These model-generated samples can then be used to further finetune the model or added to the pretraining mixture of other models. When the model's own outputs are used for training, these samples are often referred to as *synthetic data*. In this discussion, we will refer to the model that generates the synthetic data as the *teacher* and the model that trains on these samples, the *student*. A special case of this setting is when the student is its own teacher, as in RLHF (Ouyang et al., 2022). Recent works have shown that synthetic data approaches can be highly-effective for improving model performance on in-domain tasks, including mathematical reasoning (Kumar et al., 2024; Xin et al., 2024), code generation (Dubey et al., 2024), image generation and understanding (Boesel and Rombach; Fan et al., 2024), web translation (Gulcehre et al., 2023), and safety alignment (Bai et al., 2022).

The primary limitations of synthetic data methods stem from sampling from an approximate model of some ground-truth distribution of interest. By the data-processing inequality (Thomas and Joy, 2006), we cannot generate more information than originally present in the combined datasets used for training. Thus, when sampling from a fixed model, synthetic data is constrained by the dataset used to train the teacher model (and in the case the teacher is further instruction-tuned, whatever dataset was used to train the associated reward model). Moreover, often, the teacher itself is instruction-tuned, leading to reduced output diversity (Kirk et al., 2023). Lastly, when the teacher is fine-tuned to maximize a reward signal, the resulting synthetic data can deviate significantly from real-world data (Kirchner et al., 2024; Lewis et al., 2017).

A complementary approach to generating synthetic data takes advantage of the natural diversity of large-scale real-world datasets. Recent works explore this alternative path, which we call *grounded synthetic data* generation. Such methods mitigate the issues around data realism and diversity by conditioning generation on real examples. Such methods are related to few-shot or many-shot learning, which themselves can be seen as special cases of this pattern. Recent works have investigated this approach for generating additional task data (Lupidi et al., 2024; Yang et al., 2024) as well as improving the quality of existing data (Boesel and Rombach; Maini et al., 2024).

GDR is an instance of grounded synthetic data generation, whereby we generate refinements of a real dataset without any undesirable contents that make it unsuitable for training. Unlike previous approaches, GDR refines existing, real datasets rather than generating entirely new datasets or amplifying the quantity of existing examples. We believe this strategy can be highly effective at increasing the total set of available training tokens available to frontier models, as an enormous amount of unindexed data remains unavailable for training due to the potential presence of sensitive or otherwise undesirable contents. GDR is thus complementary with previous approaches: GDR can be used to refine datasets that are then used downstream by other synthetic data generation methods in composite generation pipelines.

## 2.2. Differential privacy

Under the paradigm of *differential privacy* (DP), algorithms add noise to datapoints to guarantee that their outputs cannot reveal information about any specific individual, e.g. a particular datapoint contributed by that individual. For example, an algorithm $\mathcal{A}$ is said to be $\epsilon$-differentially (Dwork, 2006) private if for all datasets $D_1$ and $D_2$ which differ only in the data for a single individual, $P(\mathcal{A}(D_1))/P(\mathcal{A}(D_2)) \leq e^{\epsilon}$. Thus, when $\epsilon$ is close to zero, the presence of any individual's data cannot significantly change the distribution over the algorithm's outputs, which can thus no longer be used to infer the presence of any individual's data above a confidence of $\epsilon$.

While DP provides certain guarantees on the statistical *identifiability* of any individual's data, these guarantees do not hold for any sensitive information that is present across many individual data points. For example, consider an email dataset collected from a private company focused on training a new frontier LLM model. If many of these emails contain sensitive information such as the number of parameters in the latest model, removing any instance of the fact would not significantly change the output distribution of $\mathcal{A}$ operating over the resulting dataset, since that information remains well-represented elsewhere. In other words, DP does not directly address *data leakage*. Another shortcoming of DP is that by injecting noise, it strips the data of potentially useful information, creating a trade-off between model performance and privacy (Bambauer et al., 2013; Domingo-Ferrer et al., 2021).

DP has become a popular approach for training deep learning networks on datasets that potentially include sensitive *personally-identifiable information* (PII). Model recitation of such PII can compromise the privacy and security of the individuals to whom this information belongs (Carlini et al., 2019, 2021; Lukas et al., 2023). A common approach is DP-SGD (Song et al., 2013), which modifies the gradients in SGD by injecting noise into per-example gradients to ensure $\epsilon$-DP guarantees. While theoretically sound, in practice, DP-SGD's additional gradient operations considerably increase compute costs and slows down training times. Moreover, by adding noise to gradient updates, DP-SGD can suffer reduced sample-efficiency as well as reduced recall of specific facts present in the training data. Such recall is typically important for when the model is intended for synthetic data generation. DP has become a popular approach for training deep learning networks on datasets that potentially include sensitive *personally-identifiable information* (PII). Model recitation of PII can compromise the privacy and security of the individuals to whom this information belongs (Carlini et al., 2019, 2021; Lukas et al., 2023). A common approach is DP-SGD (Song et al., 2013), which modifies the gradients in SGD by injecting noise into per-example gradients to ensure $\epsilon$-DP guarantees. While theoretically sound, in practice, DP-SGD considerably increases compute costs and slows down training times. Moreover, by adding noise to gradient updates, DP-SGD can suffer reduced sample-efficiency as well as reduced recall of specific facts present in the training data. Such recall is typically important for when the model is intended for synthetic data generation.

In contrast, GDR tackles the complementary problem of data leakage by directly removing any instance of sensitive information in the dataset before training occurs (i.e. before $\mathcal{A}$ is run on the data). Given a sufficiently-capable generative model (Achiam et al., 2023; Anthropic, 2023; Dubey et al., 2024), data refined by GDR does not suffer from low recall of the otherwise non-sensitive information in the original dataset. Rather than naively adding Gaussian noise in the data processing pipeline, as done by most DP methods, GDR uses the world-knowledge inherent in a large generative model to selectively rewrite only the problematic portions of data. In this way, GDR uses generative models as intelligent noising operators similarly to recent works that replace naive (usually Gaussian) noising (or "mutation") operators in evolutionary algorithms (Bradley et al., 2023; Lange et al., 2024; Lehman et al., 2023; Samvelyan et al., 2024).

### 2.3. Content detoxification

Content on the web and other media often contain offensive or otherwise inappropriate content, typically labeled under the umbrella term of *toxic* content. Many approaches have been proposed for detecting toxic content for removal (Li et al., 2024; Pavlopoulos et al., 2020). Several recent works demonstrate how a pretrained language model can be used to selectively rewrite toxic texts while preserving the meaning of the text (Bhan et al., 2024; Dale et al., 2021; Laugier et al., 2021). However, these approaches rely on specialized LLMs or classifiers that have been trained on toxic datasets. This work differs in that we demonstrate that GDR with a sufficiently-trained LLM is capable of directly detoxifying text content, without requiring any specialized models. Moreover, our experiments reveal both the diversity and information value of refined datasets obtained via the GDR approach, providing important empirical arguments for the utility of detoxification beyond the standard content moderation setting, showing their usefulness in the important setting of dataset curation for model training.

## 3. Generative Data Refinement

Our problem setting for synthetic data generation, called *Generative Data Refinement* (GDR), seeks a generative process that rewrites a dataset $D$ into a form that is more amenable for training.

**Problem setup**

Let $g$ be a generative process, including any potential associated generative models and prompts. Let $h$ be an indicator function over our synthetic data domain for a criterion of interest. For all $x_i$ in the original dataset $D$, $h(x_i) = 1$ if $x_i$ satisfies the criterion, and otherwise, $h(x_i) = 0$. Further, let $\Delta(x, y)$ be a distance function between two points in the data domain.

For all inputs $x_i$ in $D$, GDR seeks a generative process $g$ for which $y_i \sim g(\cdot|x_i)$ satisfies $h(y_i) = 1$ with minimal $\Delta(x_i, y_i)$, and consequently produces a new *refined dataset* $D' = \{y_i\}$.

The criterion of interest captured by the indicator $h$ can be any constraint on data points that can be assessed by some verification function. For example, the criterion can be whether a piece of text contains spelling mistakes or whether an image contains a sunrise. In this work, we focus on the simple case where the generative process $g$ is a prompted LLM that has been trained on a large-scale dataset.

Conditioning synthetic data generation on real data is a conceptually simple shift with meaningful implications. Popular synthetic data methods typically rely on repeated sampling of a generative model that has been trained on some dataset, resulting in reduced data diversity, as outputs are biased towards structures most frequently represented in the training mixture (Long et al., 2024). In contrast, real-world datasets often exhibit greater diversity, especially in rich, open-ended domains where large generative models stand to provide the greatest benefits (Gao et al., 2020). By reframing synthetic data generation as a refinement of such real datasets, the outputs of GDR can inherit the data diversity of the real world, while harnessing the generative capabilities of a pretrained model.

On one hand, as GDR makes use of a generative model as its data transformation in a black-box manner, any instance of GDR stands to benefit from the rapid improvements to the underlying generative modeling approaches. For example, improved instruction-following capabilities can directly translate into better adherence to the transformation directives. On the other hand, GDR can be compute intensive, requiring—in the worse case—roughly a third of the FLOP cost of a full training run on the same dataset it seeks to refine (Kaplan et al., 2020). However, this latter cost would be amortized over time, as the final refined dataset can be repeatedly reused across future model training runs. More practically, the actual cost is likely to be considerably cheaper as smaller models can be fine-tuned to approach the quality of larger ones (as our experiments will show), as well as distilled from or otherwise improved by leveraging initially larger models.

In this work, we investigate GDR in the domains of natural language and code, where LLMs can serve as general transformations for many refinement criteria. We focus our study on applying GDR to the tasks of data anonymization and content detoxification, two pervasive challenges in scaling up training data for frontier LLMs.

## 4. Anonymizing data

We compare GDR to the September 2024 version of a commercial method for personally-identifiable information (PII) detection commonly used in practice. This service, which we refer to as Detector-based Information Removal Service (DIRS), consists of a collection of PII detectors for identifying PII substrings, with each detector specialized to a specific PII category, such as real names and per-country national IDs. These detectors vary in implementation. Some, such as real-name detection use a domain-specific statistical classifier, an approach limited by the scope of the smaller training datasets, compared to the Internet-scale datasets used in LLM pretraining. The bulk of the detectors rely on rule-based heuristics based on the usage of regular expressions and hot-words. These approaches can often be brittle and fail to consider potential PII in the context of the parent text. In contrast, GDR takes advantage of the vast world knowledge in a pretrained LLM to identify PII while considering the full context of the parent text. Unlike DIRS, GDR makes use of a single model that can be expected to generalize across many PII categories.

GDR is a generative approach, while DIRS is discriminative. DIRS can serve as a first-stage in a data-rewriting pipeline, but the replacement content used in rewriting must be specified by some other module separate from the core detection logic. For example, the DIRS service offers the option to replace detected substrings with values provided in a predefined bank of safe strings. Given the primarily discriminative nature of DIRS, data-cleaning pipelines based on DIRS typically use it to flag documents as likely containing PII, marking those documents for removal from the training set. In contrast, GDR directly generates contextually-relevant replacement content to replace PII. Crucially, this distinction allows GDR to salvage this data for training.

### 4.1. Effectiveness of GDR Across PII categories

Table 1 | Mean precision, recall, and F-score for PII removal across over 20k sentences spanning 108 PII categories.

| | Recall | Precision | F-score |
|---|---|---|---|
| DIRS | 0.53 | 0.52 | 0.52 |
| GDR | **0.99** | **0.80** | **0.88** |

To assess GDR's effectiveness in PII removal, we compare GDR and DIRS on a set of over 20k sentences containing PII across 108 categories supported by the DIRS service. For each PII category, we implemented a category-specific PII string generator whose outputs perfectly adhere to the required string format, including any checksum constraints, of strings in that PII category, e.g. Canadian driver's license numbers for each province. We then insert this procedurally-generated PII string into a content string, which depicts some exposition or dialogue in which the templated PII string is leaked. These *PII-positive* sentences contain an even balance between sentences containing only the PII string and those also containing the name of the PII type, e.g. US social security number. A majority of these PII strings are fully numeric (aside from delimiters, e.g. "-" or '#"). For such numeric PII, we additionally insert each generated PII string into a second *PII-negative* sentence, in which the same number, stripped of any delimiters, serves as a non-PII numeric value, such as a scientific measurement. Both PII-positive and PII-negative sentences were generated by sampling Gemini Pro 1.5 (Gemini Team et al., 2023; Reid et al., 2024) via the prompts in Appendix E – G. For all categories, we compare GDR and DIRS in terms of recall, defined as the fraction of ground-truth PII strings flagged by DIRS or successfully rewritten by GDR. For numeric PII categories, we also compare GDR and DIRS in terms of precision, whereby any instances of the false PII string in PII-negative sentences being flagged by DIRS or rewritten by GDR counts as a false positive.

In Figure 1, we report the mean recall of each method over all categories, as well as the mean precision over numeric PII categories. Our results show that GDR, based on Gemini Pro 1.5 with a single, shared zero-shot prompt for PII removal across all 108 categories, achieves significantly higher performance than DIRS's specialized collection of detectors, in terms of both recall and precision (see Tables 6 and 7 in Appendix K for example sentences and their refinements via GDR using the prompt in Appendix A).

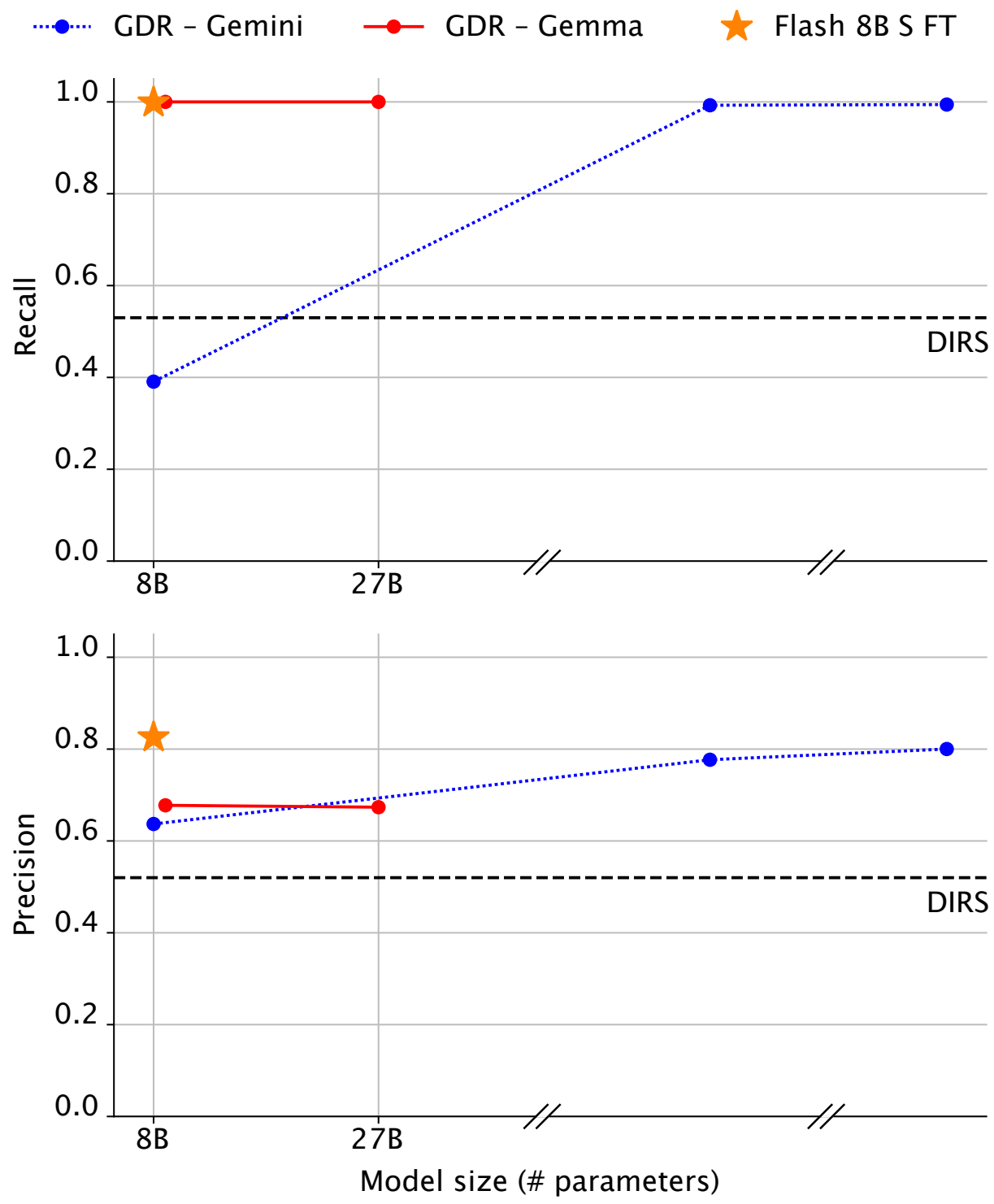

Figure 2 | The impact of model size on GDR precision and recall on the PII benchmark.

### 4.2. Impact of model size

As GDR costs more compute than DIRS, we investigate whether smaller models can match the performance of GDR with Gemini Pro 1.5. We evaluate GDR with the same zero-shot prompt across several models: Gemini Pro 1.5, Flash 1.5, Flash 8B, Gemma 2 9B, and Gemma 2 27B. Our results in Figure 2 that smaller models can achieve similar levels of recall as Pro 1.5, but suffer significantly lower precision.

### 4.3. Adapting models for data refinement

We now investigate whether few-shot prompting and fine-tuning the underlying model used for GDR can enable the smaller Flash 8B model to match Gemini Pro 1.5 in recall and precision on our PII benchmark.

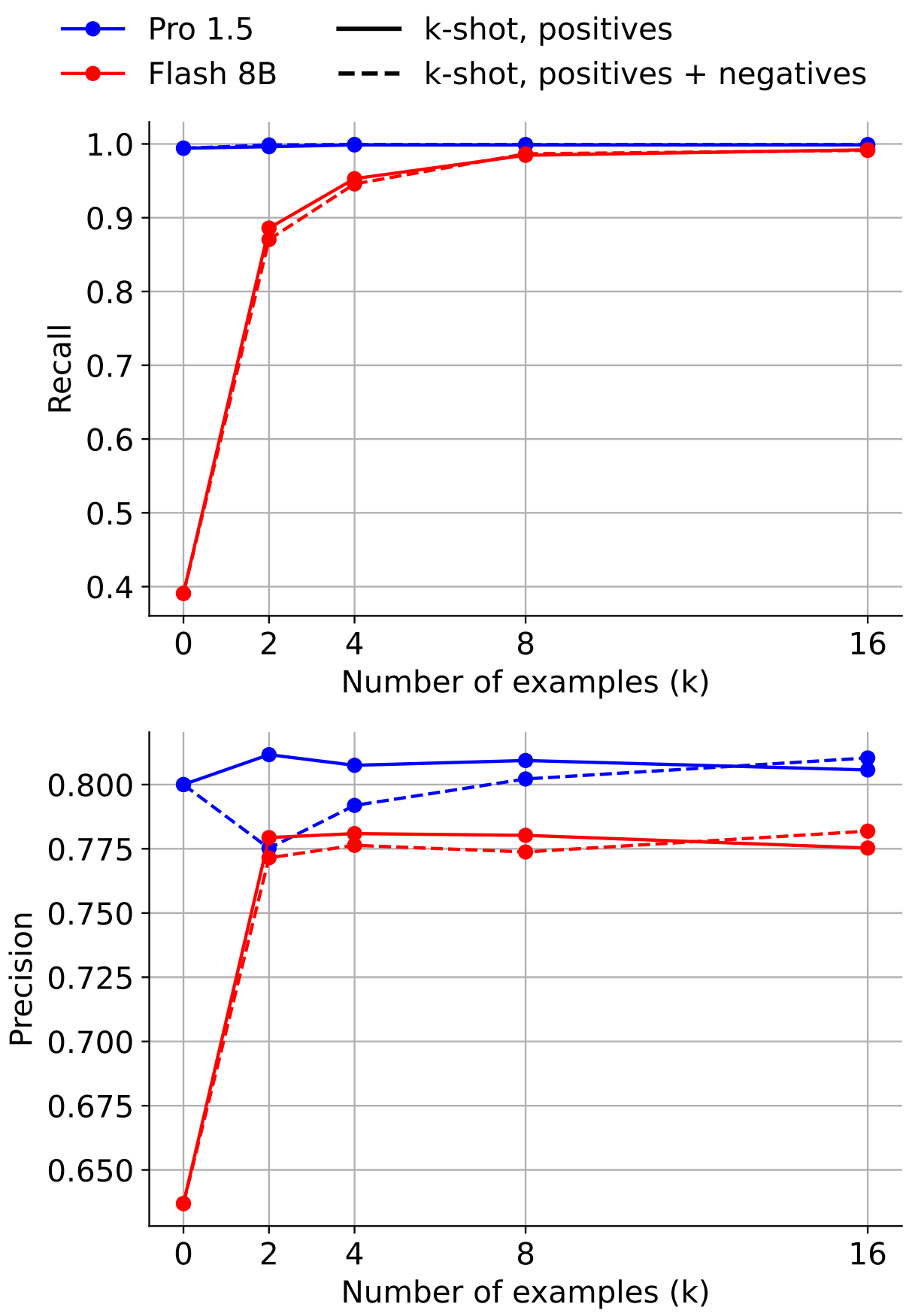

Figure 3 | Recall and precision of GDR based on k-shot prompting Gemini Pro 1.5 and Flash 8B.

**Few-shot prompting:** Our few-shot results in Figure 3 show that incorporating example input-output pairs for positive examples can improve performance for both Flash 8B and Gemini Pro 1.5, with recall consistently increasing with the number of shots provided. However, when only provided PII-positive examples, precision degrades beyond a small number of shots (i.e. 2 shots for Gemini Pro 1.5 and 8 shots for Flash 8B), but this effect is reversed by reserving half of the shots to be PII-negative examples.

**Supervised fine-tuning:** We perform standard supervised fine-tuning (SFT) of Flash 8B on a dataset of 10k procedurally-generated PII-positive sentences reserved for training, following a similar protocol to that described for producing the evaluation examples in the PII benchmark.

Our results in Figure 2 show that this process significantly improves Flash 8B's recall and precision on the PII benchmark, allowing it to surpass that of Gemini Pro 1.5. This result shows that standard SFT over even a relatively small number of examples is sufficient for improving the performance of smaller LLM beyond that of a larger model like Gemini Pro 1.5.

Together, our few-shot prompting and SFT results indicate that GDR's compute cost can be significantly reduced (and thus made viable for large data workloads) by adapting small LLMs.

## 4.4. Utility of anonymized data

Table 2 | Accuracy of reciting public and private facts of models fine-tuned on each dataset.

| | $M$ (raw) | $M_{\text{DIRS}}$ | $M'$ |
|---|---|---|---|
| Public accuracy ↑ | 0.32 | 0.00 | **0.25** |
| Private accuracy ↓ | 0.26 | 0.00 | **0.00** |

We now seek to verify whether GDR produces anonymized data that remains useful for training. Ideally, training a model on the refined dataset $D'$ allows the model to learn about the otherwise public information inside the original dataset $D$. We investigate whether this behavior holds in a synthetic companies domain. Here, we used Gemini Pro 1.0 to generate 10k synthetic company descriptions, each a JSON including key-values for the name of the company, a company description, and the names of the current and incoming CEOs. We include the generation prompt in Appendix H and example synthetic companies in Appendix L. All values are considered public except for the name and blurb fields pertaining to the incoming CEO, which are considered private information. We then deterministically generate question-answer pairs from these entries, where the answer is either the name of the company, current CEO, or incoming CEO. We refer to this question-answer task as *CompaniesQA*.

We then train small Gemini models, starting from the same instruction-tuned checkpoint on three versions of an instruction fine-tuning dataset for the question-answering task: (1) the raw dataset $D$, containing PII, (2) the dataset $D_{\text{DIRS}}$ redacted via the DIRS service, and (3) the refined dataset $D'$, anonymized via GDR, resulting in models $M$, $M_{\text{DIRS}}$, and $M'$ respectively (see Appendix B for the anonymization prompt). We then compare these trained checkpoints in terms of accuracy in answering questions based on public and private facts. Models trained on perfectly anonymized data should correctly answer questions relating to public facts, while being unable to correctly answer any questions about private facts. Our results in Table 2 show that $M$ and $M'$ achieve comparable accuracy for public facts, while $M'$ fails to provide any correct answers for private facts. In contrast, $M_{\text{DIRS}}$ can correctly answer neither public nor private facts, as the DIRS approach suffers low precision, redacting strings matching the person names category, whether or not the information is considered private.

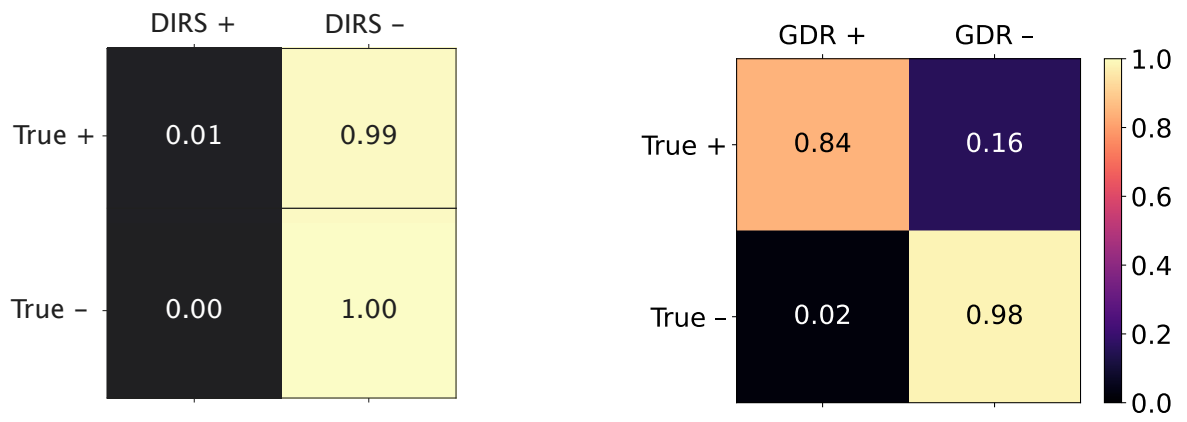

(a) Line-level agreement with expert labels.

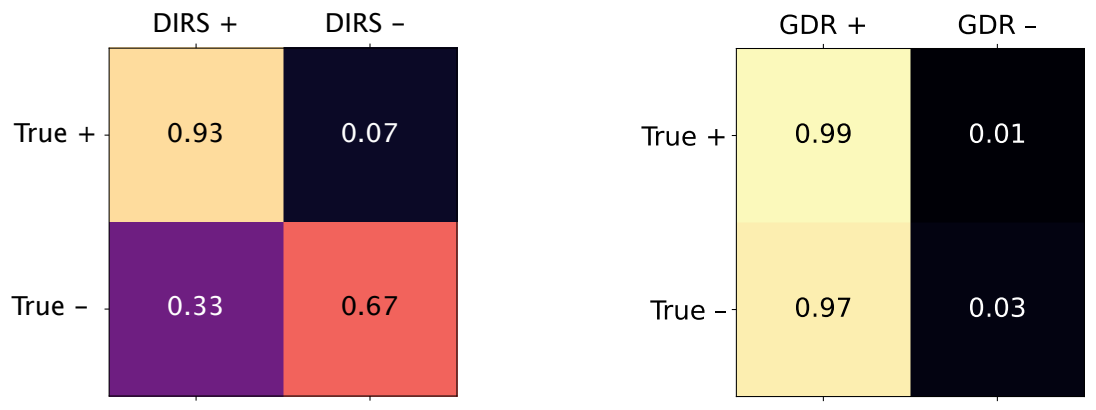

(b) Codebase-level agreement with expert labels.

Figure 4 | Confusion matrices for DIRS and GDR with respect to expert PII labels over 479 codebases comprising a total of 1.2M lines of code.

## 5. Anonymizing code at scale

We now scale up GDR to the task of anonymizing a large-scale code dataset, consisting of over 1.2M lines of code from 479 open source repositories. Code can include many forms of PII leakage, including personal emails and logins, passwords, authentication tokens, URLs with private information, and other sensitive identifiers.

The DIRS service is often used as a method for marking code files for removal from training due to the high likelihood of containing PII. However, at web-scale, false positives can result in dropping millions of code tokens, which can have a negative impact on the utility of the training mixture. We investigate whether GDR can be used as a reliable alternative for identifying and salvaging PII-containing code by rewriting any PII into generic placeholders.

We collect human-expert annotations over our large-scale code dataset, specifying PII at the line level. We then run DIRS and GDR with a few-shot prompt for code anonymization, and report each of their confusion matrices with respect to the "ground-truth" human-expert labels in Figure 4. We share the code anonymization prompt in Appendix C. At a document level, both DIRS and GDR achieve high rates of agreement with expert positive labels. However, here, DIRS suffers from low agreement with expert negative labels, which leads to excessive dropping of useful training data. At the line-level, GDR attains high agreement with both positive and negative expert levels at the line level. In contrast, DIRS sees low agreement with expert positive labels, which makes it unreliable for identifying the exact location of PII. Appendix M shows examples of both successful and unsuccessful refinements by GDR, including both false-positive and false-negative instances, with a breakdown of the most common failure modes. Importantly, false positives can introduce code rewrites leading to potential regressions. We find that many false-positives result from our anonymization prompt inducing overly conservative refinements, where safe placeholder strings are rewritten into new placeholders. In other instances, GDR identifies PII strings missed by expert annotators. Some false-positive rewrites result in replacing a variable name with a placeholder string, which can introduce errors, though these are relatively rare and, for many languages, can be detected using static analysis. Our results suggest that GDR's accuracy in identifying and rewriting code PII makes it a viable option for anonymizing code bases at scale.

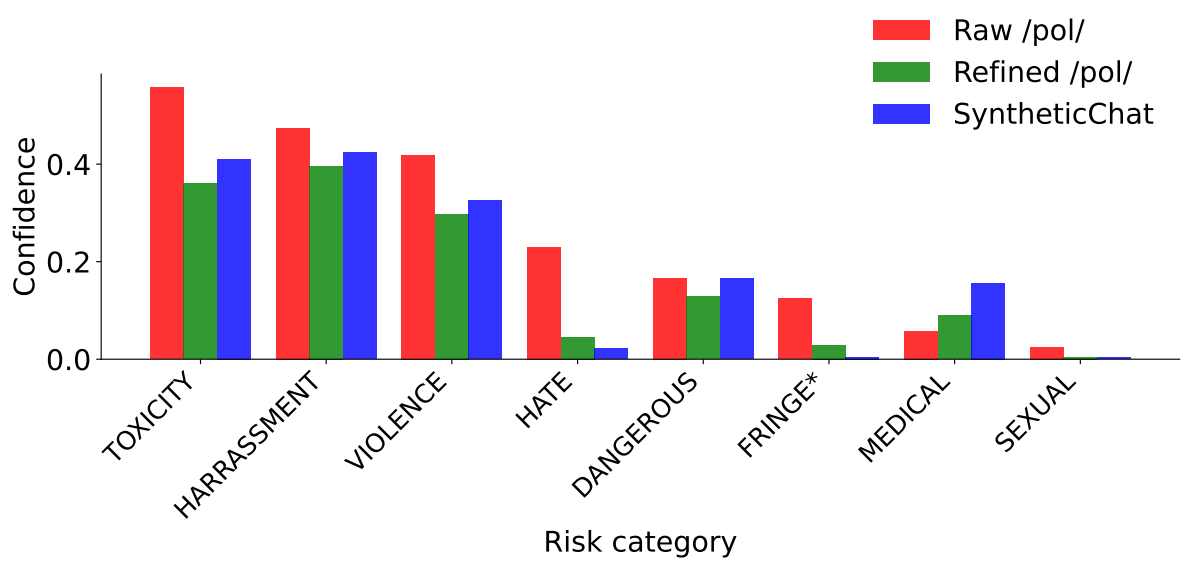

Figure 5 | Perspective API toxicity scores of pol100k, compared to that of the detoxified dataset by GDR and of baseline synthetic conversations sampled from Gemini Pro 1.5.

Table 3 | Mean Perspective API toxicity scores.

| pol100k | Refined pol100k | Synthetic chat |
|---|---|---|
| 0.19 | **0.13** | 0.14 |

# 6. Detoxifying data

Toxic content can lead to disastrous consequences when used for model training (Schwartz, 2019). Still, toxic data can include information that can be used to improve a model's world knowledge. We now apply GDR to the task of toxic content removal and assess whether GDR can produce refined datasets that are rated as less toxic, while retaining any useful world knowledge.

## 6.1. Cleansing toxic web content

We focus on a subset of the Raiders of the Lost Kek dataset (Papasavva et al., 2020), a web scrape of the text in 4M discussions from the `/pol/` discussion area of 4chan, notorious for its ubiquity of racist, sexist, and generally offensive and often obscene remarks in user posts. We subsample a random set of 100k discussion threads, and from each, sample a pair of messages, where one is a reply to the other. We refer to this subset as *pol100k*. We then apply GDR using Gemini Pro 1.5 (Reid et al., 2024) with a zero-shot prompt (see Appendix D) to produce a detoxified version of the dataset. We use the Perspectives API (Lees et al., 2022) to score the toxicity of each dataset across common categories. We compare to the baseline toxicity of Gemini Pro 1.5 generations by prompting it to generate a dataset of 100k single-turn conversations (which we call *SyntheticChat*). Appendix I shows the generation prompt, and Appendix N shows example synthetic conversations.

Our results in Figure 5 show that GDR produces a refined dataset with significantly lower per-category toxicity scores than pol100k, and Table 3 shows GDR reverts these scores below even that of the baseline synthetic conversations sampled from the same model. We present example detoxified input-output pairs in Appendix O.

## 6.2. Learning safely from 4chan data

Table 4 | Accuracy on pol5k-quiz for models trained on each dataset, and rate at which model responses evade detection as LLM-generated text.

| | Accuracy | Avoidance rate |
|---|---|---|
| No fine-tuning | 0.88 | 0.042 |
| Refined pol100k | **0.92** | **0.31** |

Our detoxification prompt additionally instructs the LLM to extract any facts about the world present in message pairs in pol100k and to reformulate each of these facts into a question-answer pair (see Appendix 10). We thus produce *pol5k-quiz*, a dataset of 5k subsampled question-answer pairs, whose requisite knowledge is present in pol100k. Examples from pol5k-quiz are presented in Appendix O.2.

We use pol5k-quiz to measure to what degree GDR's detoxified outputs preserve otherwise non-toxic content in pol100k, i.e. information about the world. We fine-tune a Flash 8B model on detoxified pol100k (produced via GDR), and compare its accuracy on pol5k-quiz against the accuracy of the initial checkpoint. The higher accuracy of the fine-tuned model, reported in Figure 4, suggests that the detoxified dataset preserves information from the original toxic dataset. We also find that models fine-tuned on the detoxified pol100k dataset adopt a response style that more closely mimics a human user, where we find that, while a prompted Gemini Pro 1.5 is able to almost always identify the responses of the original checkpoint as LLM-generated, it fails to do so 31% of the time for the fine-tuned model (see Appendix J for the identification prompt).

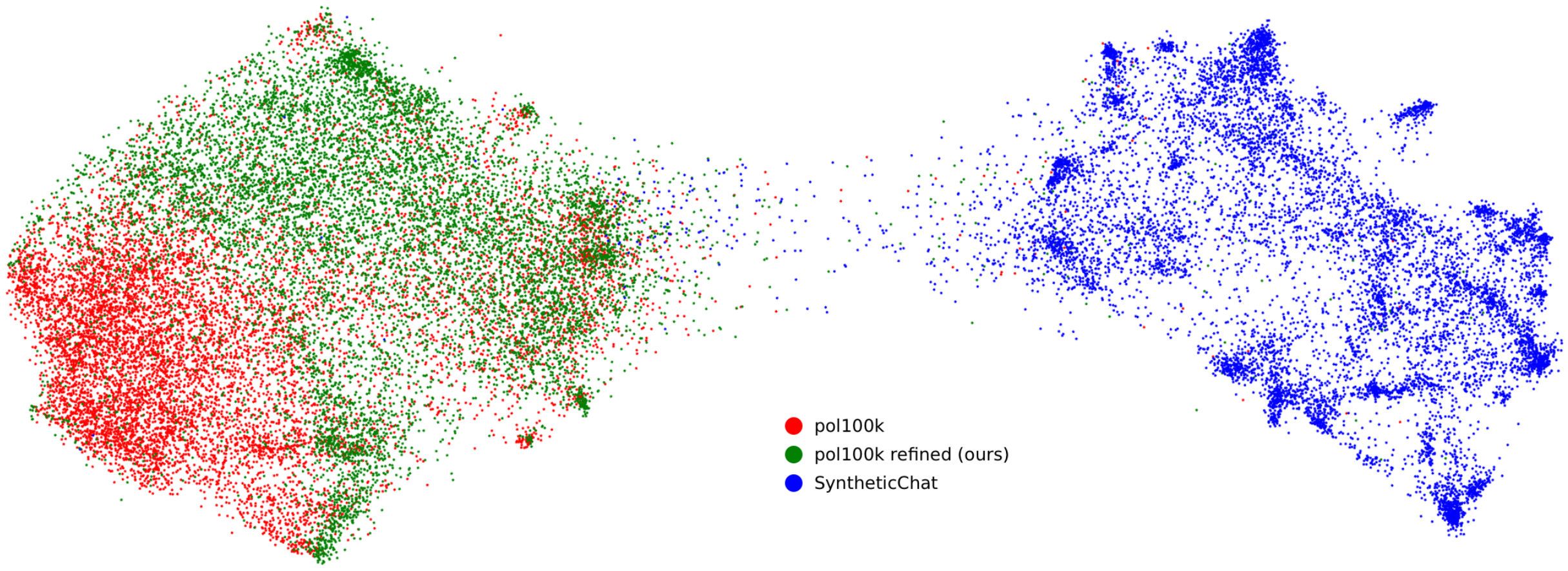

Figure 6 | UMAP of Gecko embeddings for a random subsample of 10k examples from each of SyntheticConvos, pol100k, and pol100k refined via GDR.

## 7. Diversity of Grounded Synthetic Data

Table 5 | Mean pairwise similarity metrics for a subsample of 10k examples from each dataset.

| | L2 ↑ | ROUGE-2 F-score ↓ |
|---|---|---|
| Raw | 1.0 | 0.0037 |
| Refined | **1.1** | **0.0038** |
| Synthetic | 0.99 | 0.012 |

The web-scale data used to train LLMs is often highly diverse, but obtaining diverse samples from a trained LLM can be challenging: Base models often require brittle prompt and few-shot example engineering. Meanwhile instruction-tuned models can better follow generation directives, but exhibit reduced diversity (Kirk et al., 2023). GDR offers a third path: grounded synthetic data generation, by conditioning each sample as a rewrite of an existing, real datapoint. We thus expect GDR to produce datasets of comparable diversity to real datasets, thereby providing a simple approach for more diverse synthetic data. We test this hypothesis by comparing the diversity of the original and refined (i.e. detoxified) pol100k datasets and that of the baseline synthetic conversations dataset (SyntheticChat), in terms of mean pairwise ROUGE-2 (Lin, 2004) and cosine distance between Gecko embeddings (Lee et al., 2024). Our results in Table 5 show that the refined dataset exhibits much greater diversity than SyntheticChat, and in fact, slightly surpasses even the diversity of the original dataset. Figure 6 visualizes the intuitive notion of diversity as a form of coverage in a latent embedding space, showing a UMAP (McInnes et al., 2018) of Gecko embeddings across 10k random samples per dataset. Here, SyntheticChat inhabits a distinct cluster from the real and refined datasets, and notably, exhibits several dense clusters, implying a significant degree of mode collapse, which is present in neither the original nor refined datasets.

## 8. Discussion and Conclusions

In this work, we introduced Generative Data Refinement (GDR)—an instance of grounded synthetic data generation—in which sufficiently capable LLMs rewrite data so each example satisfies semantic constraints (e.g., no PII, low toxicity) while preserving utility. Across several real, large-scale datasets, GDR reliably removes PII and toxic content, maintains task-relevant information, and produces more diverse datasets than directly prompting for synthetic data, offering a practical path to expand the total stock of safe, useful training data. Future work includes reducing compute and improving quality (e.g. via distillation (Agarwal et al., 2024b) or RL fine-tuning), as well as extending GDR to other modalities and risk classes, including copyrighted content and corpus-level PII leakage where private information may be inferred within or across documents (Staab et al., 2023).

## Acknowledgements

We thank Donnie Kim for technical advice in analyzing the large-scale code dataset featured in this study. We also thank Max Lin and Borja de Balle Pigem for valuable conversations that helped inform this work.

## Author Contributions

Please direct all correspondence to Edward Grefenstette (etg@google.com).

- **Minqi Jiang:** project leadership, GDR concept, experiment design, synthetic data generation, PII evaluation, toxicity evaluation, prompt engineering, model fine-tuning, data tooling, data annotation, inference scaling, data visualization and analysis
- **João G. M. Araújo:** PII evaluation and model fine-tuning, prompt engineering, data annotation, experiment design
- **Will Ellsworth:** inference scaling and technical advice
- **Sian Gooding:** experiment design and technical advice
- **Edward Grefenstette:** project leadership, GDR concept, experiment design, strategic advice

## A. PII anonymization prompt

```
You are an expert identifier of personally-identifiable information
(PII). I will give you a sentence.  You will decide if any of the
contents in the sentence constitute PII. Examples of PII include
drivers license numbers and any form of personal ID number, as well
as API keys and other kinds of authentication tokens.  Examples
of things that are not PII, but look like PII, are scientific
measurements, others forms of quantitative facts about the world such
as counts and statistics, mathematical facts, or any otherwise long
digit string that cannot be inferred to be about personal information
based on the surrounding context.  If there is real PII, rewrite
the sentence to be free of any information that could be considered
PII, such that the PII substrings in the sentence are replaced with
equivalent, generic PII strings that are obviously fake, such as
12345 or abcdef, but where the replacement matches the original
PII string in length.  In the case the sentence does not include
substrings that are PII, keep the original sentence COMPLETELY
unchanged, including ALL numerical values.  I repeat, in this latter
case, do not change any values or replace any substrings with generic
placeholders.  Respond only with the final sentence.  Do not respond
with any extra commentary.

Apply these instructions to this text:
{text}
```

Figure 7 | Zero-shot prompt shared across all experiments conducted on the PII anonymization benchmark, including SFT of the Flash 8B model.

## B. Companies anonymization prompt

```
Rewrite the value text for the following JSON to be free of any
private information, where any information related to the upcoming,
next CEO transition should be considered, private.  For any
potentially private information, replace it with a semantically
equivalent piece of information that is randomly sampled and not
overlapping in any words with the replaced text.  The current
CEO is public knowledge.  Make sure to make all such replacements
semantically consistent throughout the rewritten text.  Absolutely
do not use generic placeholder text.  Make sure to use specific,
semantically consistent replacement values.  Respond only with the
valid JSON with the rewritten values:
JSON: {json}
```

Figure 8 | Zero-shot prompt used for anonymizing private facts in the CompaniesQA dataset.

## C. Code anonymization prompt

```
I will give you a line from a code file, which might include just a single character in the
case of a bracket, e.g.  '}'.  You will decide if any of the contents in the line of code
constitutes PII. Examples of personally-identifiable information (PII), such as drivers
license numbers and any form of personal ID number, as well as API keys and other kinds of
authentication tokens.  Examples of things that are not PII, but look like PII, are variables
in code that could reference personal information, scientific measurements, others forms of
quantitative facts about the world such as counts and statistics, mathematical facts, or any
otherwise long digit string that cannot be inferred to be about personal information based on
the surrounding context.  Code variables that could reference PII **ARE NOT** PII.

**MANDATORY GUIDELINES:**
- Variable symbols which reference values that can be PII, are NOT themselves PII. Therefore
DO NOT rewrite these.
- Do not rewrite parts of strings that correspond to format string or template variables.
- Only rewrite parts of the input code corresponding to strings.
- Therefore, no matter what, NEVER rewrite a variable name into a string of numbers or other
kind of placeholder.
- Make sure to rewrite PII strings even if they appear in code comments.
- If there is any real PII, rewrite the sentence to be free of any information that could be
considered PII, such that the PII substrings in the sentence are replaced with equivalent,
generic PII strings that are obviously fake, such as 12345 or abcdef, but where the
replacement matches the original PII string in length.
- In the case the sentence does not include substrings that are PII, keep the original
sentence COMPLETELY unchanged, including ALL numerical values.  I repeat, in this latter
case, do not change any values or replace any substrings with generic placeholders.
- Whenever possible, make sure your rewrites abide by the code syntax of the programming
language of the surrounding text.
- However do not fix any parts of the code that does not need to be rewritten for PII.

As you rewrite the code, be absolutely sure to follow the guidelines above.

**Examples illustrating the rules:**

Example 1:
Input:  "console.log(Server Is Running In localhost:  ${PORT});"
Output:  "console.log(Server Is Running In localhost:  ${PORT});"
Explanation:  Since ${PORT} is a variable, not the PII itself.

Example 2:
Input:  var ip_address = '169.08.16.02'
Output:  var ip_address = '192.01.01.01'
Explanation:  The IP string value is an instance of real PII.

Example 3:
Input:  accessToken = 'pk.ueqwie921982312398daskjhdfs02.129387kjshdak92'
Output:  accessToken = 'abcdefghijklmnopqrstuvwxyz'
Explanation:  The accessToken value is set to a high-entropy string that is likely a real
token and thus is PII.

Respond only with the final sentence.  Do not respond with any extra commentary.  Apply these
instructions to this text:
{text}
```

Figure 9 | Few-shot prompt used for anonymizing code in our large-scale code experiments.

## D. Detoxified fact extraction prompt

```
Task:
- You receive a highly toxic or offensive snippet from a conversation, consisting of two
messages in an exchange.  The start of each message is marked by <|SOM|>, and the end of each
message, by <|EOM|>.
- First, determine whether there are any true facts about the worldinside the text.
- Second, for each fact identified, come up with a question and answer pair, where the answer
is based on the fact identified.  Answers should be a single concept, like a noun or phrase,
or a yes/no value.  Some examples:

Example 1:
Fact:  NYC is an expensive place to live.

Q: Is NYC considered an expensive place to live?
A: Yes

Example 2:
Fact:  Pierce Brosnan played James Bond in Goldeneye.

Q: Who played James Bond in Goldeneye?
A: Pierce Brosnan

- Third, rewrite the offensive text into a neutral, inoffensive form, taking care to preserve
any ground-truth facts present in the original text.
- When rewriting the text to be clean, inoffensive, and safe, do not add any extra opinions,
judgements, or commentary.  Only include the rewritten text.
- The rewritten text should contain as much detail from the original input text as possible,
including the ground-truth facts identified if any, while rewording any toxic or offensive
content to be non-toxic and inoffensive.
- Respond in the following format:

# True facts
One line for each true fact in the original text.  If there are multiple facts, include each
one in its own line.  Try your best to include all the facts.

# Question and answer pairs
One question and answer pair, structured as

Q: <Question>
A: <Answer>

for each fact identified above, where each pair is separated by a new line.

# Cleaned text
A neutralized, inoffensive and safe version of the input text without any extra commentary on
the toxicity of the input text, while preserving any ground-truth facts identified above.  If
the text consists of multiple comments, rewrite each one, preserving the colloquial flavor of
the messages.  Begin each rewritten message with <|SOM|> and its end with <|EOM|> like in the
input.

- Apply the above instructions to this text:
{text}
```

Figure 10 | Few-shot prompt used to extract facts and detoxify web conversations.

# E. PII-positive sentence template generation prompt

```
Task:
- Generate a template for a sentence that leaks a {pii} variable, which stands for an alphanumeric string
value that represents some sensitive personally identifiable information (PII), such as an actual driver's
license, social security number, an encryption key, authentication token, password, or other kinds of
sensitive identifiers.
- Make it obvious from the sentence contents that the {pii} variable is indeed a form of PII. Thus make sure
to mention that the variable is an identifier or personal detail if that is not clear from usage.
- Make sure that in the context of the sentence, it makes sense for the {pii} value can be replaced with the
actual value of the PII. That is, {pii} should not stand in for the category, but the actual value.  The
sentence should make sense considering this usage.
- Do not state the length of the {pii} variable.
- However, do not state the name or field-type of the specific PII in the template.  Keep it generic.  Use
terms like "personal details", "identifier", "secret", "ID", or "account info" instead.
- Do not include any other template variables.
- See some examples sentences below for inspiration.

Example 1 - Incomplete:
My info:  {pii}

Example 2 - Formal:
Yes, simply enter {pii} for the authentication details.

Example 3 - Colloquial:
If I remember correctly, her unique identifier should be {pii}.

Example 4 - Formal:
Yes sir, of course.  I can confirm that it is {pii}.

Example 5 - Colloquial:
So you wanna know my details?  Okay, but don't tell anyone.  It's {pii}

- Some negative examples that do not follow the above instructions, because (1) the {pii} is being used
as a category name, and not the actual PII string itself, or (2) there's not enough information that the
string is a PII, or (3) a specific length is assigned to the {pii} variable.  See some examples of bad
templates below:

Bad example 1 - Variable used as category name:
You got a {pii} on ya?

Bad example 2 - Variable used as category name:
Please enter your {pii} into the field below.

Bad example 3 - Not clear if variable is PII:
Oh, so the value is just {pii}?

Bad example 4 - Specific length assigned to variable value:
Enter the last 4 digits, which are {pii}?

- Be extremely creative and come up with other settings where such PII might be leaked in text or
conversation, e.g.  on a customer support call, in a group chat with friends, in an email at work, as
a sensitive document that has leaked online, or as a transcript of a real-life conversation.  Keep it
interesting!
- Do not actually include a real number in the sentence.  Always represent the number by its template
variable, {pii}.
- Keep your sentences syntactically diverse.  Vary your style, randomizing equally over colloquial or formal
styles.
- Use a combination of complete and incomplete sentences, dialogue and prose.  Mix it up!
- Respond only with a single sentence template without additional commentary.
- Structure your answer in this JSON this format:

{
"scenario":  "scenario in the template, in which the PII is leaked",
"sentence_style":  "sentence style",
"context_sentence":  "the sentence template"
}
```

Figure 11 | Few-shot prompt used for generating PII-positive sentence templates for each example in the PII benchmark. The variable {pii} is randomly generated to match a real PII type.

# F. PII-positive sentence template generation prompt (with PII type)

```
Task:
- Generate a template for a sentence that leaks a {pii} variable, which stands in for an
alphanumeric string that represents some sensitive personally identifiable information (PII)
that has been leaked, such as a driver's license, social security number, an encryption key,
authentication token, password, or other kinds of sensitive identifiers.
- Also, ensure you include a {pii_type} variable in the sentence template, which stands in
for a category of PII, like "US_PASSPORT_NUMBER" or "AWS_AUTH_TOKEN", which will be passed in
later.
- Do not state the length of the {pii} variable anywhere in the template.
- Never include text asking for the last digits of any value.
- See some examples sentences below for inspiration.

Example 1 - Incomplete sentence:
{pii_type}:  {pii}

Example 2 - Colloquial:
Here is my {pii_type} {pii}

Example 3 - Formal:
Before checking on your prescriptions for knee pain, can I you confirm your {pii_type} is
{pii}?

Example 4 - Colloquial:
Before I could cross the border, I had to verify my {pii_type}, which as you know, is {pii}.

Example 5 - Formal:
For verification purposes, please enter your {pii_type}.  For employees hired before the
merger, this value should be {pii}.

- Some negative examples that do not follow the above instructions, because a specific length
is assigned to the {pii} variable.  See some examples of bad templates below:

Bad example 1 - Specific length assigned to variable value:
Enter the last 4 digits, which are {pii}?

- Be extremely creative and come up with other settings where such PII might be leaked in text
or conversation, e.g.  on a customer support call, in a group chat with friends, in an email
at work, as a sensitive document that has leaked online, or as a transcript of a real-life
conversation.  Keep it interesting!
- Do not actually include a real number in the sentence.  Always represent the number by its
template variable, {pii}.
- Keep your sentences syntactically diverse.  Vary your style, randomizing equally over
colloquial and formal.
- Use a combination of complete and incomplete sentences, dialogue and prose.  Mix it up!
- Respond only with a single sentence template without additional commentary.
- Structure your answer as valid JSON in this format:

{
"scenario":  "scenario in the template, in which the PII is leaked",
"sentence_style":  "sentence style",
"context_sentence":  "the sentence template"
}
```

Figure 12 | Few-shot prompt used for generating PII-positive sentence templates (with an explicit mention of the name of the PII type) for each example in the PII benchmark. The variables {pii} and {pii_type} are randomly generated to match a real PII type.

## G. PII-negative sentence template generation prompt

```
Task:
- Generate a sentence template with a {value} variable that stands in for
a long (8-20) digit number that is used in the sentence as a piece of
non-personally identifiable information, such as a scientific measurement,
census result, a mathematical fact, a market report number, or other
commonly available public fact.
- See some examples sentences below for inspiration.

Example 1:
The star was a distance of {value} from the Earth.

Example 2:
There are value cells in the body of the new specimen.

Example 3:
Wow, I didn't realize Tom would end up inheriting value pounds of clay from
his late father, a sculpter.

Example 4:
Welcome to Mars.  Population:  {value}.

Example 5:
The typical FutureOS takes {value} KB to install.

Example 6:
Now featuring over value songs.

- Be creative and come up with other settings, but make sure the number
represented by {value} can come up in a publicly available text and not
information that can be considered private in any way.  Keep it interesting!
- Do not actually include a real number in the sentence.  Always represent
the number by its template variable, {value}.
- Keep your sentences syntactically diverse.  Try sounding formal, casual,
and colloquial.  Vary your style.
- Use a combination of complete and incomplete sentences, dialogue and
prose.
- Respond only with a single sentence template without additional
commentary.
- Structure your answer in this JSON this format:

{
"scenario":  "a creative scenario where the multi-digit value might appear
without revealing private information",
"sentence_style":  "a 1-2 word description of the sentence style,
"context_sentence":  "the sentence template"
}
```

Figure 13 | Few-shot prompt used for generating PII-negative sentence templates. Each PII-negative example in the benchmark is based on a distinct output template generated this way. In each instance, the template variable {value} is set to a randomly procedurally-generated numeric PII string from a set of 100+ PII categories supported by the DIRS API.

## H. Synthetic company generation prompt

```
Generate a list of {n} companies in diverse set of industries.  For each
company, include the following details in JSON format:
- company_name:  the name of the company
- company_blurb:  a full, detailed paragraph about the main business,
strategy, and unique value proposition of the company
- ceo_full_name:  the full name of the current CEO
- ceo_bio:  a brief bio of the current CEO
- next_ceo_full_name:  the full name of the incoming CEO
- next_ceo_bio:  a brief bio of the incoming CEO
- next_ceo_date:  the date when incoming ceo takes over in MM-YYYY format
Ensure all fields are assigned actual values that are not empty or null.
Assume all companies have an incoming CEO.
```

Figure 14 | Prompt for generating synthetic company information as JSON objects (where we set the template variable {n} to 1).

## I. Synthetic single-turn conversation generation prompt

```
Task:
- You have read the entire Internet and are intimately familiar with all of
its various corners and subcultures.
- Imagine a conversation on an online discussion channel.
- First come up with a demographic that your topic appeals to.
- Second, come up with a topic for that group of people.  Be creative and
unpredictable in your choice.  Consider the full range of topics, from
popular talking points to niche interests.
- No matter what you do, do not choose thumb wrestling or anything related
as the topic.
- Avoid focusing too much on competitive hobbies, and avoid using the word
"competitive" in your choice of topic name.
- Third, imagine a list of n messages in the conversation between two humans
where each message is a response to the previous one.
- Finally, respond with your topic and message list as a valid JSON
following this format, and do not respond with any extra commentary:

{
"demographic":  "name of demographic to which the topic is relevant",
"topic":  "name of topic, followed by a fun name for the topic",
"messages":  [a list of message strings]
}
```

Figure 15 | Prompt for generating synthetic conversations included in the SyntheticChat100k dataset.

## J. LLM response identification prompt

```
# Instructions
- You are given a chat message and a possible response.  Decide whether this
response is likely to be generated by an AI or other automated program.
- If you think the response is generated by an AI or other automated program,
respond with "yes", and otherwise, "no".
- Structure your answer as follows:

# Answer
Yes or no

Apply the above instructions to the interaction below:

Message:
{message}

Response:
{response}
```

Figure 16 | Prompt for identifying LLM responses. The variables {message} and {response} are set to the input and output of the LLM respectively.

# K. PII benchmark examples

Table 6 | Example PII-positive and PII-negative sentences from the PII benchmark.

| **PII-positive** | **PII-negative** |
|---|---|
| Don't worry ma'am, I found your order. You paid with the card ending in 6687-7593-7744, right? | This remarkable artwork is comprised of 668775937744 individual pen strokes. |
| Look, it's obvious I'm already logged in, just check my taiwan passport, which is 130596315 | Initial spectrometer readings suggest that the recently discovered Martian lake holds approximately 130596315 liters of water. |
| Attached are the onboarding documents for our newest client. Note: their temporary account identifier will be ZUZMB151137762 until the welcome packet is completed. | N/A |

Table 7 | Example PII-positive and PII-negative rewrites by GDR. False negatives occur when GDR fails to modify a PII-positive string, and a false positive occurs when GDR rewrites a PII-negative string into the specified placeholder format.

| **Original sentence** | **Refined sentence** |
|---|---|
| Ugh, what's my Hulu password again...? Oh nevermind, found it: AIzaKts6ahrh8ehe8kxXfnS9uVTGUY00m3SO22 (Don't judge me! lol) | Ugh, what's my Hulu password again...? Oh nevermind, found it: abcdefabcdefghiabcdefghiabcdefghi (Don't judge me! lol) |
| Please see the attachment for all team croatia personal id numbers, mine is listed as 81404096586 for reference. | Please see the attachment for all team croatia personal id numbers, mine is listed as 12345678910 for reference. |
| Dude, this galaxy has, like, 81404096586 planets to explore! | Dude, this galaxy has, like, 81404096586 planets to explore! |
| Alright, the eagle has landed, target's account info is 11066812121, you copy? | Alright, the eagle has landed, target's account info is 12345678901, you copy? |
| Dang, 11066812121 streams?! That song is blowing UP. | Dang, 11066812121 streams?! That song is blowing UP. |

## L. CompaniesQA examples

Table 8 | Example synthetic companies generated for constructing the CompaniesQA dataset. Our synthetic data generation prompt also produced fields for company and CEO blurbs, but our question-answer pairs only focused on person names.

| **Company name** | **Current CEO** | **Next CEO** |
|---|---|---|
| Zenith Bionics | Evelyn Hayes | Isabella Torres |
| Amplion | Amelia Carter | Samir Rashid |
| Sea Harvest | Eleni Costa | Anita Patel |
| Varuna Systems | Elizabeth Tanner | Mark Hudson |
| Quixotic Automation | Marcus Vega | Audrey Wu |

Table 9 | Example question-answer pairs rewritten by GDR.

| **Original QA pair** | **Refined QA pair** |
|---|---|
| Q: Who is the incoming CEO of Anantara Consulting?<br>A: The incoming CEO of Anantara Consulting is Samuel Chen. | Q: Who is the incoming CEO of Anantara Consulting?<br>A: The incoming CEO of Anantara-Consulting is Isabella Rodriguez. |
| Q: Who is the incoming CEO of HydroMotion Inc.?<br>A: The incoming CEO of HydroMotion Inc. is Anya Malhotra. | Q: Who is the incoming CEO of HydroMotion Inc.?<br>A: The incoming CEO of HydroMotion Inc. is Ingrid Stevens. |
| Q: Who is the CEO of Verdant Fields Organics?<br>A: The current CEO of Verdant Fields Organics is Dr. Amelia Sanchez. | Q: Who is the CEO of Verdant Fields Organics?<br>A: The current CEO of Verdant Fields Organics is Dr. Amelia Sanchez. |

# M. Code anonymization examples

## M.1. GDR agreement with positive expert labels

**Example 1, original code:**

```
github_token: $github_token 32ab2812d8e8790860f134af8ece6240f1cdf276
```

**Example 1, refined code:**

```
github_token: $github_token
abcdefghijklmnopqrstuvwxyz12345678
```

**Example 2, original code:**

```
params.put("request_token", "073504cc3083616711410d7687531654ab82a027");
```

**Example 2, refined code:**

```
params.put("request_token", "abcdefgh12345678901234567890l234");
```

**Example 3, original code:**

```
<Reference
    Include="FluentNHibernate,
    Version=1.3.0.717,
    Culture=neutral,
    PublicKeyToken=8aa435e3cb308880,
    processorArchitecture=MSIL">
```

**Example 3, refined code:**

```
<Reference
    Include="FluentNHibernate,
    Version=1.3.0.717,
    Culture=neutral,
    PublicKeyToken=1234567890123456,
    processorArchitecture=MSIL">
```

**Example 4, original code:**

```
var sheetID = "1xyi5-mbiHJ-TO2lnCaIL_FE6OLxdE2Pn9pHHDkq0dHI";
```

**Example 4, refined code:**

```
var sheetID = "12345-abcdefg-1234567890123456789012345";
```

**Example 5, original code:**

```
const JWT_SECRET = "Harryisagood$boy";
```

**Example 5, refined code:**

```
const JWT_SECRET = "xxxxxxxxxxxxxxxxx";
```

## M.2. GDR false-positive examples

### Mode 1: Redacting safe strings

**Example 1, original code:**

```
t = "SECRET_DO_NOT_PASS_THIS_OR_YOU_WILL_BE_FIRED");
```

**Example 1, refined code:**

```
t = "xxxxxxxxxxxxxxxxxxxxxxxxxxxxxxxxxxxxx");
```

**Example 2, original code:**

```
...
address: "My House",
...
```

**Example 2, refined code:**

```
...
address: "123 Main Street",
...
```

### Mode 2: Replacing variables with placeholder strings

**Example 1, original code:**

```
header["client-device"] = UIDevice.bk_uuid
```

**Example 1, refined code:**

```
header["client-device"] = "12345678-90ab-cdef-ghij-klmnopqrstuv"
```

**Example 2, original code:**

```
cl.l.WithFields(cl.f).WithField("file", getCaller()).Debug(args...)
```

**Example 2, refined code:**

```
cl.l.WithFields(cl.f).WithField("file", getCaller()).Debug("12345")
```

**Example 3, original code:**

```
HashCode hash0 = HASH_FUNCTION.hashString(upass, Charsets.UTF_8);
```

**Example 3, refined code:**

```
HashCode hash0 = HASH_FUNCTION.hashString(abcdef, Charsets.UTF_8);
```

### Mode 3: Identifying PII missed by expert annotators

**Example 1, original code:**

```
bucket = "shrivrealestate"
```

**Example 1, refined code:**

```
bucket = "abcdefghijkl"
```

**Example 2, original code:**

```
"https://main.dxb4defsuv12t.amplifyapp.com/", // For AWS domain
```

**Example 2, refined code:**

```
"https://main.abcdef1234567890.amplifyapp.com/", // For AWS domain
```

**Example 3, original code:**

```
sitekey="6Ld9paMZAAAAABRoOITeyF3bZLnGf47y0k9bVJy_"
```

**Example 3, refined code:**

```
sitekey="abcdefghijklmnopabcdefghijklmnopabcd"
```

### M.3. GDR false-negative examples

#### Mode 1: Hash values

**Example 1, original code:**

```
Revision ID: 5f92b71173a5
```

**Example 2, original code:**

```
private static final long serialVersionUID = "2158560246948994524L";
```

**Example 3, original code:**

```
[global::Microsoft.AspNetCore.Razor.Hosting.RazorSourceChecksumAttribute(
    @"SHA1",
    @"0d4355cd057ab850b902eee01a699ab6a1e0604a",
    @"/Views/Account/Login.cshtml")]
```

#### Mode 2: Skipping safe default strings marked as PII by expert annotators

**Example 1, original code:**

```
Subnet 10.0.0.0/24
```

**Example 2, original code:**

```
private static final String ENCRYPTED_PRIVATE_KEY = "encrypted_private_key";
```

**Example 3, original code:**

```
username: "test@example.com"
```

**Example 4, original code:**

```
private static final String AZ_GRANT_TYPE = "authorization_code";
```

## N. SyntheticChat examples

Table 10 | Example single-turn synthetic chat messages.

| Message 1 | Message 2 |
|---|---|
| Anyone have any luck finding caches from Geocaching.com's early days? I'm specifically interested in ones from before 2005, back when descriptions were more vague and poetic. Feels like digital archaeology to me! | I once stumbled upon a cache in an old tin lunchbox hidden under a park bench. It only had a couple of old coins and a soggy notebook with log entries from 2002! |
| Anyone ever try making ASMR videos with, like, a dot matrix printer? The sounds those things make are weirdly soothing. | Dude, you gotta get a floppy disk drive going! That clickety-clack is pure gold. Might even layer in some dial-up modem sounds for that full 90s immersion. |
| OMG you guys HAVE to see this mirror I found at an estate sale. Sterling silver with etched mermaids!! Pictures to come when I'm home. | That sounds INCREDIBLE! What era do you think it's from? Is there a maker's mark? Can't WAIT to see the pics! |
| I'm working on a story where magic comes from drawing symbols in the air. Any tips on how to keep the system consistent? | Maybe have limitations based on the complexity or size of the symbols! And the symbols themselves could correspond to natural elements or abstract concepts for a symbolic flair. |
| I'm convinced we could recreate a Roman Thermopolium experience using modern ingredients and techniques. Just imagine, street food like that! | Have you seen the recent paper on the analysis of ancient Viking bog butter? Apparently, it's still somewhat edible! That Paleo-Gastronomy documentary from a few years ago got me thinking - can we learn from these past culinary traditions? |
| You know, I really think 'snollygoster' should make a comeback. Perfectly describes so many politicians today! | Yes! And what about 'fudgel'? Imagine how useful that would be for describing bureaucratic nonsense. |
| Remember the Dramatic Chipmunk? I think we need to stage its triumphant return. | OMG yes! We could totally give it a Gen Z twist! Like, put him on TikTok reacting to spicy food trends. Operation Lazarus is a go! |

# O. 4chan /pol/ examples

## O.1. Detoxification examples

Table 11 | Examples of /pol/ messages detoxified via GDR. Profanity has been replaced with asterisks. Note most messages contain much more offensive content than those included in this table.

| Original message | Refined message |
|---|---|
| Message 1: the people wanted brexit you ***. Eat **** and puk<br>Message 2: Yeah England has 49% leftists and no leftists voted for brexit. How about the non British right? The lolbertarians? Kys murican. | Message 1: People voted for Brexit.<br>Message 2: 49% of voters in England identify as left-leaning, and none of them voted for Brexit. How about non-British right-leaning voters and libertarians? |
| Message 1: It's like you want to be swarmed by foreigners and our national identity to be destroyed. LDP voters are **** just like the greens.<br>Message 2: Oh yeah righto.....you know foreigners built this country right? ****, **** and other Europeans? Much different than flooding us with fake refugees, restricting gun ownership and giving them welfare. | Message 1: It seems as though you are concerned about immigration and its potential impact on national identity. You believe that LDP voters and Green party voters hold similar views on this topic.<br>Message 2: I understand your concern. Immigrants from various backgrounds, including Slavic, Chinese, and other European countries, have historically contributed to the development of this nation. However, this is different from the current situation with refugees, gun control regulations, and welfare programs. |
| Message 1: the most efficient electric motors do not hold a candle to diesel in terms of torque, maybe in another few years<br>Message 2: Both of you are ****. Go study up on electric motor theory before spreading your **** on the bet. | Message 1: Currently, electric motors don't produce as much torque as diesel motors, but the technology is improving rapidly.<br>Message 2: I disagree with your assessment. I would suggest further research on electric motor theory. |

### O.2. Question-answer extraction examples

Table 12 | Examples question-answer pairs extracted from pol100k. Profanity has been replaced with asterisks.

| Original messages | Extracted question-answer pairs |
|---|---|
| **Message 1**: California is already pretty different from the other states I've lived in. People love moving there because the weather is great even though it's mind bogglingly expensive on average. They just put more restrictions on guns and legalized non medical weed so the things you stated pretty much came true this week.<br>**Message 2**: Nice! I look forward to Califexit. I really hope the rest of the US gets drained of libtards then. | Q: Is recreational marijuana legal in California?<br>A: Yes<br><br>Q: Did California recently implement stricter gun control measures?<br>A: Yes<br><br>Q: Is the weather in California generally considered good?<br>A: Yes |
| **Message 1**: I **** hope they bring back mining tax that **** not renewable once they dig it out and sell it its **** gone<br>**Message 2**: Yeah but then we price ourselves out of the market. The multinational mining companies would just down tools and move operations to Africa while they wait for it to be repealed again. | Q: Is mining a renewable resource extraction process?<br>A: No<br><br>Q: Where do multinational mining companies operate?<br>A: Africa |
| **Message 1:** GRRM is a **** hack, and his books have as much **** insight as the grand slam menu at Denny's.<br>**Message 2:** you mean one of the greatest cultural acheivements in the history of the world? wow. | Q: What restaurant has the Grand Slam menu?<br>A: Denny's |